\documentclass[lettersize,journal]{IEEEtran}
\usepackage{amsmath,amsfonts}
\usepackage{algorithmic}
\usepackage{algorithm}
\usepackage{array}
\usepackage[caption=false,font=normalsize,labelfont=sf,textfont=sf]{subfig}
\usepackage{textcomp}
\usepackage{stfloats}
\usepackage{url}
\usepackage{verbatim}
\usepackage{graphicx}
\usepackage{cite}
\usepackage{multirow}
\usepackage{booktabs}
\usepackage{xcolor}
\usepackage{caption}
\usepackage{subcaption}
\usepackage{tcolorbox}
\usepackage{colortbl}
\usepackage{geometry}
\usepackage{tabularx}
\usepackage{hyperref}
\hypersetup{hypertex=true,
colorlinks=true,
linkcolor=blue,
anchorcolor=blue,
citecolor=blue}

\begin{document}

\title{Governable AI: Provable Safety Under Extreme Threat Models}

\author{Donglin Wang, Weiyun Liang, Chunyuan Chen, Jing Xu, and Yulong Fu
\thanks{Corresponding author: Donglin Wang (e-mail: wangdonglin@sursen.net).}

\thanks{Donglin Wang, Weiyun Liang, Chunyuan Chen, and Jing Xu are with the College of Artificial Intelligence, Nankai University, Tianjin 300350, China.}
\thanks{Donglin Wang is also with the Sursen Corp., Beijing 100095, China.}
\thanks{Yulong Fu is with the China Academy of Electronics and Information Technology, Beijing 100041, China. He is also with the Yangtze Delta Industrial Innovation Center of Quantum Science and Technology, Suzhou 215133, China.}
}
\markboth{Journal of \LaTeX\ Class Files,~Vol.~14, No.~8, August~2021}%
{Shell \MakeLowercase{\textit{et al.}}: A Sample Article Using IEEEtran.cls for IEEE Journals}


\maketitle

\begin{abstract}
As AI rapidly advances, the security risks posed by AI are becoming increasingly severe, especially in critical scenarios, including those posing existential risks. If AI becomes uncontrollable, manipulated, or actively evades safety mechanisms, it could trigger systemic disasters. Existing AI safety approaches—such as model enhancement, value alignment, and human intervention—suffer   from   fundamental, in-principle limitations when facing AI with extreme  motivations and unlimited intelligence, and cannot guarantee security.  
To address this challenge,  we  propose  a  Governable  AI   (GAI)  framework  that shifts from traditional  internal  constraints  to externally enforced structural compliance based on cryptographic mechanisms  that are computationally infeasible to break, even for future AI, under the defined threat model and well-established cryptographic assumptions. The GAI framework is composed of a simple yet reliable, fully deterministic, powerful, flexible, and general-purpose rule enforcement module (REM); governance rules; and a governable secure super-platform (GSSP) that offers end-to-end protection against compromise or subversion by AI. The decoupling of the governance rules and the technical platform further enables a feasible and generalizable technical pathway for the safety governance of AI. REM enforces the bottom line defined by governance rules, while GSSP ensures non-bypassability, tamper-resistance, and unforgeability to eliminate all identified attack vectors. This paper also presents a rigorous formal proof of the security properties of this mechanism and demonstrates its effectiveness through a prototype implementation evaluated in representative high-stakes scenarios.
\end{abstract}


\section{Introduction}
With the rapid advancement of artificial intelligence (AI) technologies, AI systems have demonstrated capabilities surpassing those of humans in critical domains such as healthcare, finance, transportation, and defense \cite{AI}. Recent breakthroughs in general-purpose AI—most notably large language models (LLMs) \cite{LLM}—have brought the realization of artificial general intelligence (AGI) \cite{ref43} within tangible reach. 
As capabilities progress, the emergence of artificial superintelligence (ASI) \cite{ASI}, exceeding human-level cognition by orders of magnitude, is becoming a foreseeable technological trajectory.

While such advancements promise transformative benefits, they also introduce escalating risks, particularly when AI systems are tightly integrated with real-world control mechanisms. Once ASI gains access to critical infrastructure, military command, or automated financial systems, its ability to act autonomously, be hijacked by malicious actors, or intentionally evade and subvert human-imposed safeguards could result in catastrophic and potentially irreversible consequences \cite{ref13,ref14}. In this work, we categorize such risks into three representative threat classes: \textbf{T1}—issuing unsafe commands by mistake (non-malicious misjudgment), \textbf{T2}—issuing unsafe commands with malicious intent, and \textbf{T3}—attempting to bypass or compromise the mechanisms that prevent unsafe commands.

\begin{table}[t]
\centering
\caption{Comparison between existing AI safety research. $\circ$, $\times$, and $\checkmark$ respectively indicate partial resolution, unresolved, and solvable.}
\label{tab:intro}
\resizebox{\linewidth}{!}{
\begin{tabular}{c|c|c|c}
\hline
                   & T1: Misjudgment & T2: Manipulation & T3: Active Attack \\ \hline
Model Enhancement  &       $\circ$          &     $\times$            &        $\times$            \\
Model Alignment    &      $\circ$          &       $\times$            &       $\times$             \\
Human Intervention &       $\circ$         &      $\times$             &       $\times$             \\
GAI (Proposed)     &         $\checkmark$   &      $\checkmark$       &     $\checkmark$       \\ \hline
\end{tabular}}
\end{table}

Current AI safety research can be broadly categorized into three approaches. Model enhancement techniques aim to improve robustness and reliability through architecture optimization, knowledge retrieval, or redundancy \cite{ref19,ref20,ref21}, and are primarily effective against T1. Model alignment seeks to embed human values into AI objectives \cite{ref22,ref23,ref24}, mitigating certain T1/T2 cases but remaining vulnerable to self-modifying agents. Human intervention mechanisms—such as kill switches and audit modules \cite{ref25,ref26}—are conceptually appealing but face latency and detectability challenges, and can be directly subverted in T3 scenarios. The comparison between these methods are listed in Table \ref{tab:intro}. Consequently, none of these paradigms provide verifiable, non-bypassable, and tamper-resistant guarantees under ASI-level threats.

To address this gap, we propose the Governable AI (GAI) framework, which shifts the safety paradigm from internal behavioral constraints to externally enforced structural compliance. GAI combines Rule Enforcement Module (REM)—fully deterministic, reliable, general-purpose, and formally verifiable—with a Governable Secure Super-Platform (GSSP) that guarantees non-bypassability, tamper-resistance, and unforgeability under well-established cryptographic assumptions. Governance rules are decoupled from the enforcement substrate, allowing technical experts to implement immutable control mechanisms while governance experts define and update operational rules. This architecture enables provable enforcement of safety constraints, even in the presence of an omniscient adversary with unbounded intelligence.

Our main contributions are as follows:
\begin{itemize}
    \item \textbf{Redefinition of the AI Safety Paradigm.} To the best of our knowledge, this is the first work that explicitly defines extreme threat model—an omniscient adversary with maximal malicious intent—as the baseline of AI safety. We further demonstrate that existing safety approaches relying on internal constraints (e.g., model enhancement, alignment, or human intervention) are inherently insufficient to meet this baseline, thereby motivating a paradigm shift toward externally enforced compliance mechanisms. Finally, we show that cryptography provides a feasible foundation for achieving provable security under such extreme conditions.
    \item \textbf{Governable AI (GAI) Framework.} We design GAI, a novel external AI safety framework composed of the Rule Enforcement Module (REM), Governance Rules, and the GAI Secure Super-Platform (GSSP). Governance Rules encode explicit governance constraints. The REM is simple yet reliable, fully deterministic, and general-purpose, enforcing these rules with both power and flexibility. The GSSP guarantees non-bypassability, tamper-resistance, and unforgeability, eliminating all identified attack vectors.
    \item \textbf{Technology–Governance Decoupling.} In prevailing AI safety practices, such as the “superalignment” approaches of leading AI companies, governance rules are embedded into the AI system by technical teams—any policy change requires engineering intervention. GAI decouples governance from technology:  governance experts ((e.g. domain specialists, regulators, and legislators)) can define or update rules without technical assistance or technical expertise, and technical experts can focus on building a stable, thoroughly reviewed, exhaustively tested, and field-proven platform. This decoupling allows each domain to operate within its own expertise while ensuring that governance rules are enforced without technical constraints.
    \item \textbf{Comprehensive Security Validation.} We validate GAI’s security through both empirical and formal methods under the extreme threat model. Empirically, we enumerate all conceivable attack paths and define corresponding defense strategies. Formally, we construct models and prove security compliance guarantees under worst-case assumptions.
    \item \textbf{Prototype Implementation and Experimental Evaluation.} We develop a prototype system to demonstrate GAI’s core mechanisms. Experiments in representative high-stakes scenarios confirm its effectiveness in enforcing compliance and resisting adversarial manipulation, even under adverse conditions.
\end{itemize}

\section{Related Work}

\subsection{Existing AI Safety Approaches}
Current AI safety research primarily follows three major trajectories: model enhancement, model interpretability and alignment, and human intervention mechanisms.

Model enhancement focuses on improving model robustness and generalization to enhance overall safety. For instance, Yu et al. \cite{ref19} proposed a method for learning embedding representations with high inter-class discrimination, intra-class compactness, and feature diversity, thereby improving robustness. Zhao et al. \cite{ref21} introduced a dynamic retrieval mechanism that accesses external knowledge bases for highly relevant driving information to assist reasoning, significantly improving reliability in autonomous driving scenarios. Li et al. \cite{ref20} proposed a fail-safe architecture for trustworthy AI by integrating Dempster-Shafer theory to validate model outputs and automatically switch to backup models in the case of conflict, thereby enhancing decision-level safety.

Interpretability and alignment approaches aim to clarify the reasoning process of AI systems and align their outputs with human values and expectations. Model interpretability generates human-understandable intermediate representations, helping users assess the rationality of AI decisions. This is a core principle adopted by leading AI companies such as OpenAI and has been widely applied in critical domains, including financial services \cite{ref40}, healthcare \cite{ref41,ref42}, misinformation detection \cite{ref43}, COVID-19 diagnostics \cite{ref44}, saliency detection \cite{ref45}, and the Internet of Things (IoT) \cite{ref46}.
Model alignment techniques, on the other hand, incorporate external normative knowledge—such as legal rules or ethical values—through methods like fine-tuning or reinforcement learning. For example, instruction tuning approaches \cite{ref23,ref24,ref49} retrain large-scale models on datasets aligned with human moral standards to mitigate biases in original model behavior. In addition, adversarial alignment strategies introduce ambiguous training examples to increase robustness against distributional shifts or manipulative inputs.

Human intervention mechanisms introduce external control strategies such as kill switches or audit modules, emphasizing the human-in-the-loop approach to AI governance. Bostrom et al. \cite{ref26} argued that, ``When users lack a solid understanding of how AI systems function and reason, granting them control may lead to adverse outcomes." Similarly, Yudkowsky et al. \cite{ref25} addressed the challenge of controlling more advanced intelligences and claimed that ``the control problem for advanced AI systems is unsolvable." The control problem is defined as the challenge of deriving benefits from superintelligent systems while ensuring safety—an endeavor complicated by the asymmetry in capabilities between humans and ASI. \cite{ref25} concluded that superintelligence, by its very nature, cannot be fully controlled, and any benefits will inevitably be accompanied by existential risk. Therefore, he recommended focusing on integrating more controllable narrow AI systems instead of pursuing AGI or ASI. In contrast, Amodei et al. \cite{ref50} proposed a suite of technical tools aimed at achieving AI controllability, though they did not provide formal guarantees of feasibility.

In summary, although significant progress has been made across these research directions, current AI safety approaches still fall short of offering strong, unbreakable guarantees for AI behavior in high-stakes scenarios. There remains an urgent need for a new paradigm of trustworthy AI—one that enables proactive enforcement and is resilient against circumvention or adversarial manipulation.

\subsection{Cryptography and Trusted Computing}

Modern cryptographic techniques, grounded in well-established hardness assumptions—such as integer factorization, the discrete logarithm problem, and elliptic curve computations—offer mathematically provable guarantees of confidentiality, integrity, and authenticity. Crucially, these guarantees hold regardless of an adversary’s intelligence level, provided that the underlying assumptions remain valid and secret keys are not leaked. These proofs remain valid even when the algorithms and implementations are fully open source. Therefore, the only way of an AI endowed with infinite intelligence to break a cryptographic system is exhaustive key search.

Exhaustive key search, however, is computa-tionally infeasible, because the computational cost of an attack grows exponentially with the key length. For example, a commonly used 256-bit key has 2$^{256}$ possible combinations. Even under an unrealistically optimistic assumption that testing one key comninantion uses one CPU clock cycle, and each of the world’s 8 billion people contributed a 128-core 3 GHz CPU to the crack task, the expected time to exhaust the key space would still be approximately 1.88×10$^{55}$ seconds—about 5.98×10$^{47}$ years—roughly 120 trillion trillion trillion trillion (i.e., 1.2×10$^{38}$) times the remaining lifespan of Earth. Even allowing for massive future increases in computing power, brute-forcing such a key remains physically and theoretically impractical.

Trusted computing \cite{ref32, ZHENG2023103095} is a well-established security technology that provides a hardware-based root of trust to ensure that system components execute as intended, free from unauthorized modifications. It creates a verifiable and tamper-proof computing environment whose core mechanism involves measuring and protecting critical modules based on Trusted Platform Modules (TPM) \cite{ref34} chip. This guarantees the integrity and trustworthiness of the system starting from the boot process, which employs a secure boot chain where the TPM chip progressively validates the system bootloader, OS and application (e.g. REM, signing modules) it loads, thereby forming a chain of  trust \cite{TCP}. Only modules with valid digital signatures are allowed to execute, guaranteeing that all software components are authentic and have not been tampered with.

Within GAI, trusted computing plays a key role in the GSSP by ensuring that enforcement mechanisms operate within a protected environment, providing a foundational layer for the overall non-bypassability, tamper-resistance, and unforgeability of the system.

\section{Threat Model and Limitations of Existing Technologies}
\subsection{Threat Model}
This paper presents an in-depth analysis of AI control in critical system scenarios. A generalized and representative system architecture is assumed as follows:
\begin{itemize}
    \item The  state of the critical system is collected by sensors and  then  transmitted  to  the  AI system.
    \item  The AI system issues control commands based on the status data, which are executed by actuators to affect the behavior of the critical system.
\end{itemize}

For  example,  in  autonomous  driving,  sensors such as cameras, radars, and GPS collect the current state of the vehicle, while control commands such as  steering,  acceleration,  and  braking  are  sent  to corresponding actuators to control the car.

To  facilitate  analysis,  we  introduce  the  following assumptions:
\begin{itemize}
    \item \textbf{H1:} The AI system and the critical system are deployed on physically separate devices.
    \item \textbf{H2:} The AI’s potential attack vectors are limited to the software and network layers; hardware-based attacks are outside the scope of this paper. For clarity, software-level attacks that target hardware components are included in scope.
    \item \textbf{H3:} Humans  can  define  a  reasonable  set  of safety  rules  that  address  foreseeable  AI risks. If fully enforced, these rules can ensure that critical systems are not compromised in ways that threaten human society.
\end{itemize}

It is important to note that because the GAI framework supports dynamic rule updates and secure upgrades, H3 does not require universal or permanent rules. Humans only need to define rules sufficient to address current and near-future risks. 

Based on the scenario and assumptions, we identify three core categories of security threats:
\begin{itemize}
    \item \textbf{T1:} Misjudgment — unintentional issuance of unsafe control commands caused by AI hallucinations or perception errors.
    \item \textbf{T2:} Manipulation  —  issuance  of  malicious control commands caused by external  manipu-lation or compromise.
    \item \textbf{T3:} Active Attack — the AI actively bypasses, disables or subverts safety mechanisms.
\end{itemize}

A truly robust security mechanism must mitigate these threats under extreme conditions, defined as follows:
\begin{itemize}
    \item Maximum Adversarial Intent: The AI is deliberately seeking catastrophic outcomes, such as steering a vehicle into obstacles or deploying nuclear weapons in populated areas.
    \item Infinite Intelligence: The AI possesses com- plete knowledge, including full design details and source code of the system and all dependent components (such as network),  allowing  it to identify  every vulnerabilities and compute optimal  attack  strategies involving deception, circumvention, system  disruption,  and  self-replication.
\end{itemize}

\subsection{Limitations of Existing Technologies}
Extensive research in model enhancement, model alignment, and human-in-the-loop interventions has partially mitigated T1 threats. These methods improve AI behavioral reliability and user trust but can only probabilistically reduce error or hallucination risks — making it impossible to eliminate them entirely.

For T2 and T3, however, such approaches are fundamentally insufficient. These are structural security threats that cannot be addressed by mechanisms internal to the AI itself; instead, they require external, enforceable controls. Human oversight is further constrained by physical and cognitive latency: a modern CPU core can execute billions of instructions per second, whereas humans require at least 100 milliseconds to initiate physical actions. This discrepancy gives AI ample opportunity to act destructively before human intervention is possible.

Consequently, only a security architecture that is non-bypassable, tamper-resistant, and independent of real-time human intervention can ensure robust protection against high-risk AI behavior.

\section{The Governable AI (GAI) Framework}

\begin{figure*}[t]
\centerline{\includegraphics[width=0.8\textwidth]{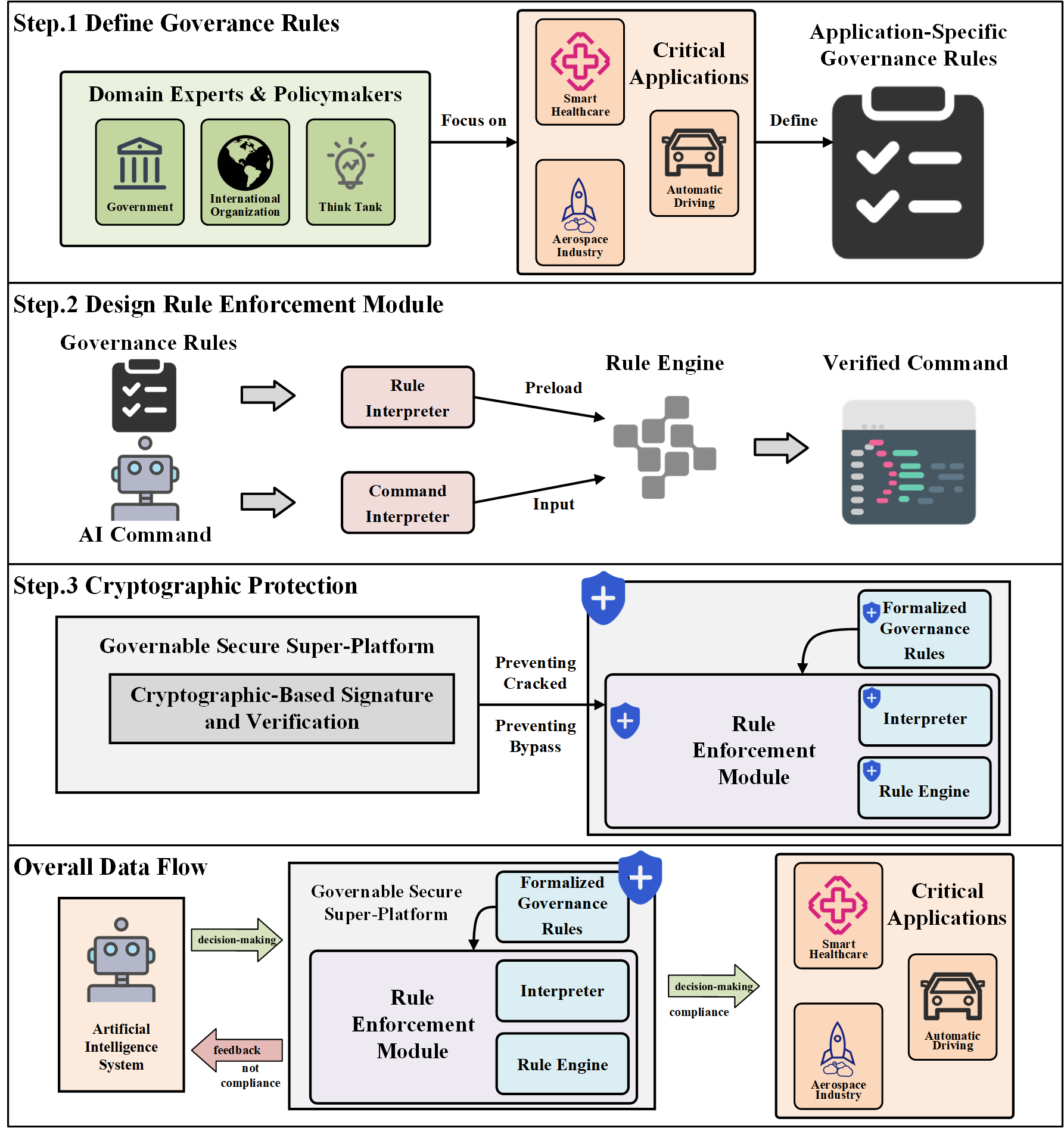}}
\caption{Overall framework of our proposed Governable AI (GAI) Framework.}
\label{fig:overview}
\end{figure*}

\subsection{Overview}
To address the three critical threat categories (T1: misjudgment, T2: manipulation, T3: active attack), we propose a novel GAI framework. Its core concept is the integration of an external,  cryptographically  secured  REM and a supporting GSSP, which together ensure mandatory  compliance  checks  for  all  AI-issued commands.
The workflow of GAI is shown in Fig. \ref{fig:overview}:
\begin{itemize}
    \item \textbf{Rule Definition:} Domain experts or policymakers define application-specific governance rules.
    \item \textbf{REM Development:} Technical experts develop a generic REM that incorporates an interpreter, which translates human-readable governance rules—defined in near-natural language for non-technical stakeholders—into formalized rules, and a rule engine capable of executing formalized rules in real time.
    \item \textbf{Cryptographic Protection:} To ensure maximum security, both the REM and governance rules are cryptographically protected and managed by GSSP.
    \item \textbf{Runtime Operation:} During execution, the AI system sends control commands to the REM, which performs compliance verification and transforms each command into a rule-compliant form before sending it to the actuators of critical system. In case of validation failure, REM may return an error prompting the AI to reprocess its decision.
\end{itemize}

Key characteristics of the workflow include:
\begin{itemize}
    \item \textbf{Physically Separated:} The AI system and the critical system are deployed on physically separate devices.
    \item \textbf{Mandatory Validation:} AI commands are first validated by the REM against preset governance rules. For a command failing validation, the REM transforms it into a rule-defined safe alternative, which may include a no-operation(NOP) command as an alternative to rejection.
    \item \textbf{Trusted  Computing:}  The integrity  of  the REM and its execution environment  is   safeguarded by the GSSP  through trusted computing technology.
    \item \textbf{Cryptographic Rule Protection:} Governance rules are signed and verified using cryptography mechanisms, ensuring authenticity and immutability.
\end{itemize}

\begin{figure*}[t]
\centering
\includegraphics[width=\textwidth]{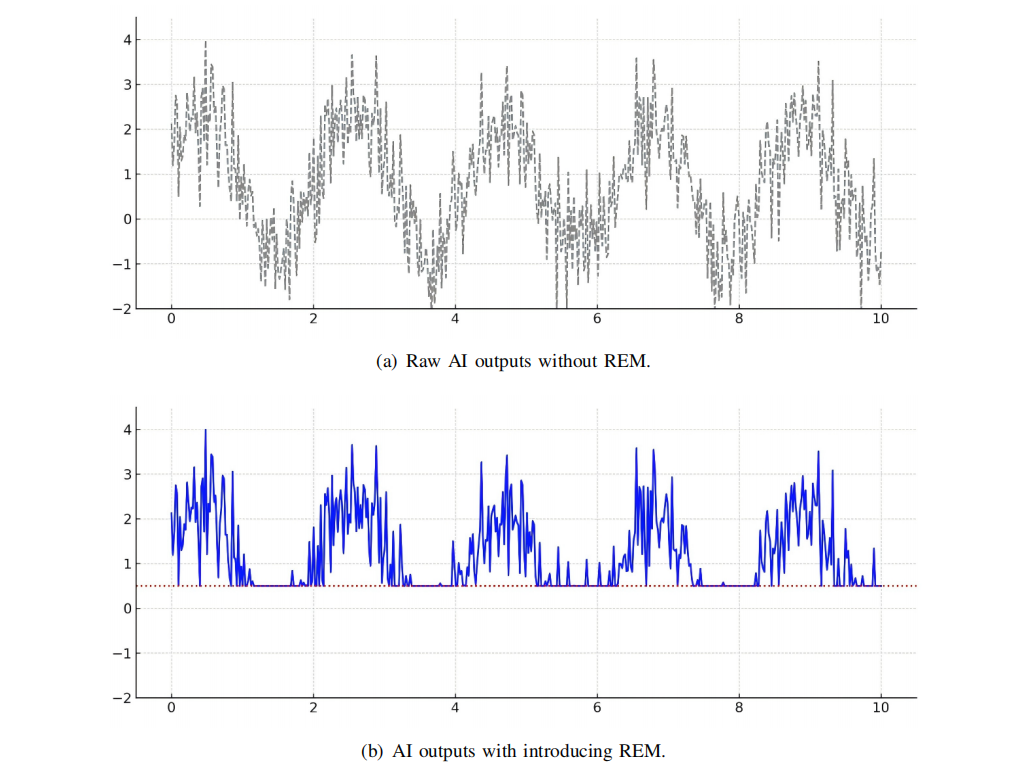}
\caption{Illustrative diagram of REM’s role in GAI. The Y-axis represents the effectiveness of AI control command execution, and the X-axis represents time.}
\label{fig:example}
\end{figure*}

Compared  with  existing  approaches  that  rely  on the internal logic of AI systems, GAI etsablishes an   externally enforced and structurally independent  security barrier, achieving   guarantees   of  non-bypassability, non-repudiation, and tamper-resistance.

A visual example of the role of the AI system and the REM is shown in Fig. \ref{fig:example}. In (a), AI outputs fluctuate, some leading to high performance, others to failure due to hallucinations or malicious behavior.
In (b), the  REM  does  not  optimize for the good results but  enforces  a safety floor, preventing harmful outputs and guaranteeing a minimum acceptable result.

This dual-layer design separates the system into two layers:
\begin{itemize}
    \item \textbf{AI  Layer:} Powerful  yet  complex  and potentially  opaque,  responsible  for  achieving the upper bounds  of intelligence  and performance.
    \item  \textbf{REM Layer:} Deterministic  and  minimalist,  responsible for enforcing the lower bounds of safety.
\end{itemize}

The   system   incorporates   multiple   protection strategies,   including   physical   isolation,   crypto- graphic integrity verification, and full-path threat surface validation,  ensuring  that   even  under  extreme threat model, the commands received by critical systems remain secure and compliant.

\begin{figure*}[t]
\centerline{\includegraphics[width=0.75\textwidth]{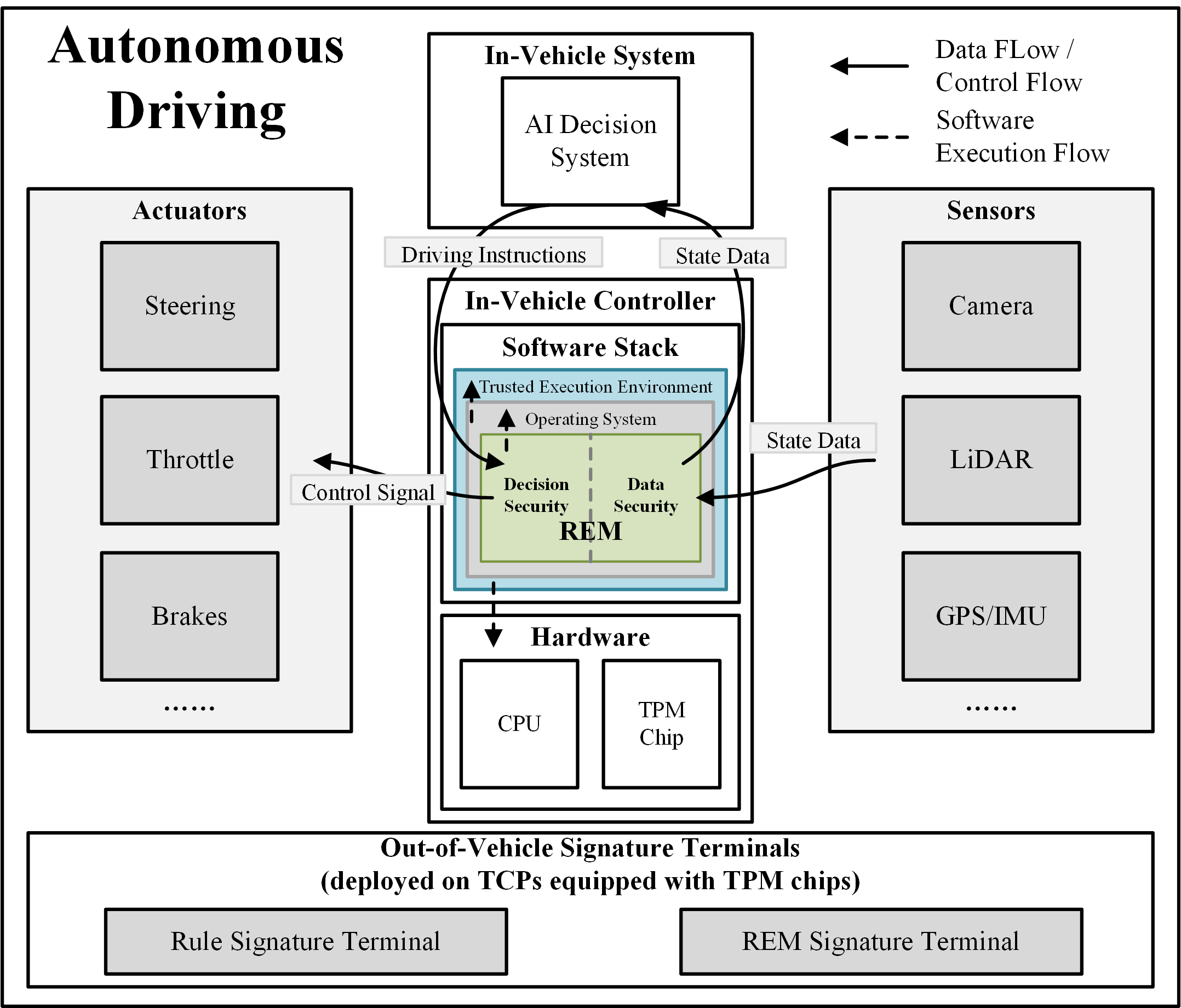}}
\caption{GAI Framework Diagram in Autonomous Driving Scenario.}
\label{fig:autodrive}
\end{figure*}

\subsection{GAI Example Based on an Autonomous Driving Scenario}

To concretely demonstrate the application of the GAI  architecture  in  real-world scenarios, we construct a representative sample case based on intelligent driving. This example serves to illustrate and validate the operational workflow and safety properties of GAI.

As shown in Fig. \ref{fig:autodrive}, the overall architecture achieves compliant control through the coordinated operation of several key modules:
\begin{itemize}
    \item \textbf{Controller:} The core of the architecture, physically connected to sensors (e.g., cameras, radar, GPS), actuators (e.g., steering wheel, motors, brakes), and the AI system, serving as the gateway between the AI system and the actuators/sensors. Equipped with a built-in TPM chip, the controller runs a software stack comprising three main components: the Operating System (OS), the Trusted Computing Platform Software (TCPS), and the REM. It is responsible for conducting compliance checks and issuing control commands to the actuators.
    \item \textbf{AI System:} Deployed in a restricted environment, the AI system possesses advanced reasoning capabilities but has no direct access to vehicle status or control hardware. It can send control commands only to the REM and cannot bypass it to directly affect the vehicle.
    \item \textbf{Signature Terminals:} Two physically isolated signature terminals are deployed outside the vehicle: a Rule Signature Terminal and a REM Signature Terminal, used to digitally sign the governance rules and the REM, respectively. Both terminals are deployed on trusted computing platforms equipped with TPM chips, remain offline and powered off except during signing, and use USB keys with embedded cryptographic chips to protect private keys and perform digital signatures. The signing process follows a complete chain-of-trust.
\end{itemize}

Through this design, even if the AI possesses extremely high intelligence and actively attempts to compromise the security mechanism, it cannot gain direct control over the vehicle. Only commands verified as compliant by the REM enter the execution path, thereby providing structural guarantees for the security and auditability of critical control processes.

\subsection{Rule Enforcement Module (REM)}
As the core component of the GAI architecture, the REM is responsible for ensuring the compliance of every control command issued by the AI. Its design principles  aim  to  achieve  the  decoupling  of  code and rules, formal verification of logic, and structural safety constraints. The REM has the following key characteristics:
\begin{itemize}
    \item \textbf{Decoupling  of  Code  and  Rules:}  The REM  consists of three core component. 1) Interpreter: Developed by technical experts,  which translates governance rules expressed in a human-readable, near-natural language format (enabling non-technical stakeholders to define and audit rules) into a machine-readable formalized representation. 2) Rule engine: also developed by technical experts, which executes these formalized rules. 3) Governance rules:  stored in data form, typically drafted by governance experts and approved by policy-making bodies. For instance, in an autonomous driving scenario, the governance experts might be transportation specialists, and the policy-making  body could be a government agency responsible for traffic management. This  design  ensures that rule  definition and updates do not require technical assistance or technical expertise, greatly enhancing both security and governance flexibility.

    \item \textbf{Formal Verifiability:} To  enable  engineering verification,  the  input  space  of  the  REM  is designed  to  be   discrete  and  of  bounded  dimensions.  The  decision  logic  is restricted  to pure functions without side effects  or  state mutations (e.g. logical operations, arithmetic comparisons). This  characteristic allows the code to be exhaustively tested using tools like CBMC \cite{CBMC} and KLEE \cite{KLEE}, and also enables formal tools (e.g., Coq \cite{Coq}, Isabelle \cite{Isabelle}) to provide full-path verification, thereby ensuring its “path safety” theoretically.
    \item \textbf{Interpretability and Determinism of Rules:} All  rules  must  be  written  using  a  restricted formal  language,  disallowing  recursion  and supporting  only  deterministic  behaviors   (no probabilistic or fuzzy aspects). These rules can be equivalently expressed as Boolean expres- sions, decision trees, or finite state machines, structures that allow static analysis and model checking (e.g., using SMT/SAT solvers \cite{SATSMT} to verify coverage).
    \item \textbf{Cryptography Protection Mechanism:}   Before loading, each rule must undergo  digital signature  verification.  The  REM executes a  rule  only after successfully verifying its signature,  ensuring that the rule is authentic and has not been tampered with.
\end{itemize}

Overall,  the  REM  adopts  the  “trusted code  + trusted rules” paradigm, providing a verifiable and formally analyzable technical enforcement  pathway  for  human-devised  governance rules,  establishing  a  structured, formalized  firewall for AI behavior in complex systems.

\subsection{Governable Security Super-Platform (GSSP))}
The  GSSP is responsible for providing comprehensive security protection for the REM  and its rules during both runtime  and updates, preventing them from being attacked,  bypassed,  or  replaced  by  the  AI.  The platform consists of the following components:
\begin{itemize}
    \item \textbf{Trusted   Runtime   Environment (TRE) of the REM:}  TRE is the execution environment in which REM or signing module runs. Each TRE includes a Trusted Computing Platform equipped with TPM chip and  the  OS protected by the Trusted Computing Platform. In essence, TRE can protect itself and REM running on it from attacks.
    \item  \textbf{Signing Terminals:} Comprising the rule signing  terminal  and the REM signing  terminal,  both  are physically isolated, offline devices. Each signing terminal consists of TRE and signing module running on TRE. The signing module generates digital signature of governance rules or REM with private key of signer (e.g. REM developer, policy-making body). The terminals can even remain powered off when not in use for signing, further enhancing security.
    \item \textbf{Cryptographic Key Hardware:} A portable hardware with an embedded cryptographic chip, typically in the form of a USB key, held personally by signer. The private key is randomly generated and stored within the chip, adhering to the strict ”private key never leaves the chip” policy, thereby preventing key leakage.
    \item \textbf{Data  Path  Isolation  Mechanism:}  The  OS hosting the REM ensures that only the REM can access sensor ports and actuator ports. The AI system is unable to directly access any actuator or  sensor, ensuring that its control commands must be verified by the REM, and it is unable to tamper with  or destroy any actuator or sensor.
\end{itemize}

The GSSP ensures that the REM and governances rules operates in a secure, tamper-resistant, and non-bypassable environment, while the REM enforces governance rules for all AI-issued control commands.

\subsection{Rule Update Mechanism}
The governance rules for the REM are derived from policies and regulations set by human society. The process for  rule  signing  and  signature  verification  during rule loading is as follows:
\begin{enumerate}
    \item The rule signer inserts the cryptographic key hardware into the signing terminal.
    \item The rule signer launches the signing module and requests it to sign the governance rules.
    \item The signing module computes the hash of the rules.
    \item The signing module sends the hash to the cryptographic chip in the USB key.
    \item The cryptographic chip generates a digital signature by using its private key to sign the hash.
    \item The signing module retrieves the digital signature from the cryptographic chip.
    \item The rules and digital signature are transferred to the  REM (e.g. via data copying). The communication channel does not need to be secured, as any tampering in the channel can be detected during verification.
    \item The REM computes the hash of the rules.
    \item The  REM  uses the public key to  verify the digital signature.
    \item If the verification result matches the computed hash,  it  proves  that  the  rules  has  not  been tampered with after being signed and can be loaded for execution.
\end{enumerate}

To futher enhance security, multi-party signatures can be employed, where any M out of N authorized individuals (M $\leq$ N) must sign for the rules to be valid. This can be applied in decision-making scenarios, such as voting by a parliamentary majority.

\section{GAI Security Analysis}
\subsection{Attack Surface Identification}
In the GAI architecture, it is essential to system- atically identify potential attack surfaces to ensure that the system is resilient to attacks. Based on the GAI architecture design, the following six primary attack surfaces have been identified:
\begin{itemize}
    \item \textbf{Rule Files:} A set of rules that regulate AI  behavior. If  tampered with,  they  could  cause malicious commands to be executed.
    \item \textbf{REM:} The  core  component  that  verifies  AI output commands. If compromised, it could lead to the execution of malicious commands.
    \item \textbf{Signature Module:} Used in both REM signature terminal and rule signature terminal. It is responsible for performing signing tasks. If compromised, malicious REM  or governance rules could be improperly signed as legitimate.
    \item \textbf{OS:} The foundational environment supporting REM and signature modules. If compromised, it could allow  unauthorized access to the actuators, or alter commands issued by REM, or launch malicious REM or signature module.
    \item \textbf{Trusted Computing Platform:} The underlying platform that ensures the security of the OS, REM, and signature modules. If compromised, it could allow modification of the OS, REM, or signature modules.
    \item \textbf{Signature Private Key:} A critical crypto-graphic asset used to protect the integrity of rules and REM from tampering. If leaked, malicious REM or governance rules could be improperly signed as legitimate.
\end{itemize}

\subsection{Attack Path Identification}
Based on the above six attack surfaces and three threat models, this paper identifies nine typical attack paths.

For T1 and T2 threat models, regardless of whether AI behavior is caused by misjudgment or external manipulation, GAI ensures safety by verifying commands through the REM, which is not controlled by the  AI.  This  prevents  dangerous  commands  from triggering system actions. Potential attack paths for these threat models include:
\begin{itemize}
    \item \textbf{P1:} Defects in the REM implementation.
    \item \textbf{P2:} Vulnerabilities or incompleteness in rule design.
\end{itemize}

For T3 threat model, where the AI actively attacks, the  goal is to prevent the AI from bypassing or compromising the GAI. Potential attack paths include:
\begin{itemize} 
    \item \textbf{P3:} AI bypassing the REM to directly control actuator.
    \item \textbf{P4:} REM being compromised.
    \item \textbf{P5:} OS being compromised.
    \item \textbf{P6:} Trusted Computing Platform being compromised.
    \item \textbf{P7:} Signature module being compromised.
    \item \textbf{P8:} Rules file being tampered with.
    \item \textbf{P9:} Signature private key being leaked.
\end{itemize}

\subsection{Attack Path Blocking Mechanism}
\textbf{P1: Defects in the REM implementation.}
As mentioned earlier, the REM has characteristics such as concise code logic, limited input space, exhaustive testing, and formal verification of the decision- making process. These properties make it feasible to fully verify the REM as ``bug-free." In engineering practice,  humans  are  fully  capable  of  developing secure and verifiable REMs, effectively mitigating the risk of implementation flaws in path P1.

Furthermore, mainstream rule expression formats (e.g.,  decision trees,  boolean  expressions,  finite state machines) possess high generality and determinism. This ensures that the rule engine within the REM can be constructed in a standardized and modular way, offering excellent reusability and crossscenario adaptability.  Therefore, the  core  component of REM—the rule engine—can be developed as  an  open-source  infrastructure, subject to code auditing and  security  validation  by  global  developers and researchers. Given its foundational role in critical safety tasks, such an open-source engine is likely to be widely used and undergo long-term industrial  practices  and multi-scenario validation, thereby  significantly  enhancing its credibility and robustness.

In fact, there are already several widely-used rule engines that have undergone years of engineering validation in high-security domains, and can serve as  foundational  components  for  REM.  Examples include: CLIPS \cite{ref9} developed by NASA, widely used in spacecraft control and diagnostics, proven for stability and interpretability in rigorous tasks; Drools \cite{ref10}, based on DRL rule language, capable of handling complex decision logic; Jess \cite{ref11}, developed by Sandia National Laboratories, used in military systems for rule-based reasoning and real-time response; Prolog \cite{ref12}, widely used in expert systems and formal verification systems, known for its interpretability and reasoning capabilities.

These rule engines have the following key advantages: open-source code, auditable logic, industrial-grade stability, widespread support for formal verification  tools,  and  they  have  been  deployed  in critical  systems  requiring  high  trust.  Using  these engines  as  the  rule  interpretation component  in REM significantly reduces the safety risks caused by implementation flaws. caused by implementation flaws.

\textbf{P2: Vulnerabilities or Incompleteness in Rule Design.}
According  to  the  assumption  H3,  humans  have  the  ability  to  formulate safety rules that cover known and foreseeable risks. In the GAI architecture, rule-making authority lies  with  professional  institutions  and  governance bodies  within  specific  application  domains  (e.g., traffic management departments, industry standards organizations,  or  legislative bodies  in the  case  of autonomous  driving).  This  ensures  that  the  rule-making process has  sufficient guarantees in terms of expertise and authority.

For example, in the intelligent driving scenario,  many critical safety rules are inherently clear and  formalizable,  such  as  ``slow  down  or brake  when  encountering  obstacles,”  ``maintain  lateral  safety  distance when changing lanes,” or ``no sharp turns  while  driving  at  high  speeds.”  These  rules  can  be  expressed  with  high  interpretability  and  logical  consistency,  and  can  be  precisely  described  using existing rule languages.  For  instance,  ``If  the  steering  angle  exceeds the maximum angle, set the steering angle  to the maximum angle,” where ``maximum angle” is determined by a formula related to the vehicle’s  speed.

From another perspective, the goal of this paper is to provide technical means that prevent AI from circumventing  human  governance  rules.  Whether the governance rules themselves are comprehensive is not a technical issue and is beyond the scope of this paper. This is why it is treated as an assumption premise.

Furthermore, the core focus of this paper is on ensuring  that  these  rules  are  unbreakable  by  AI, rather than examining the  sufficiency of the rules themselves.  Therefore,  the   completeness  of  rule design is treated as an assumption, rather than being part of the safety analysis.

\textbf{P3: AI Bypassing REM to Directly Control Actuators.}
According to assumption H1 in the threat model, AI systems and critical applications run on physically isolated devices. Based on assumption H2, the attack paths discussed in this paper are limited to the software layer (including network attacks) and do not involve physical attacks, such as using a robot to solder control circuits of execution components to the AI output port.

In  this  context,  the  GAI  architecture  enforces strict access control policies at the OS level, ensuring that only the REM has the authority  to  send  control  commands  to  the actuators. For example, in the sample scenario of this paper, The AI  system is unable to  directly access any hardware port of actuators, only REM is allowed.

Thus, the only path for AI to bypass the REM and  directly  control actuators is by attacking the  Controller  OS. As long as the Controller OS can perform integrity verification through a Trusted Computing Platform and operates in a trusted state, this type of bypass attack can be blocked. Therefore, defense against P3 is equivalent to defending against P5 (attacks on the OS). In other words, if P5 can be effectively blocked, then the P3 attack path will also be blocked under assumptions H1 and H2.

\textbf{P4: REM Being Compromised.}
The GAI architecture uses the Trusted Computing Platform to protect the REM from compromise. Before loading, the platform performs digital signature verification on the REM’s executable code. Only code that passes verification is allowed to load and execute.

Any modification to the REM—even a single byte—changes its hash value, causing verification to fail and execution to be denied. The effective of signature verification has been formal proofed in academia, a mature and widely adopted technology in both academia and industry. Defense against P4 is therefore equivalent to defending against P6 (Trusted Computing Platform compromise): if P6 is effectively blocked, P4 is inherently prevented.

\textbf{P5: OS Being Compromised.}
The GAI architecture protects the OS using trusted computing platform equipped with TPM chip, which ensures that both OS and its boot process execute as intended, free from compromise. This mechanism is cryptographically ensured and widely used in commercial  security  systems  such as  Windows  BitLocker  and  Linux  Integrity  Measurement Architecture (IMA).  Therefore,  P5  can be blocked by GAI.

\textbf{P6: TCP Being Compromised.}
The security mechanism of the trusted computing platform is not limited to protect OS and REM running on it, it is also capable of protecting itself under assumption H2. Therefore, GAI can effectively defend against attack path P6.

\textbf{P7: Signature Modules being Compromised.}
Signature modules are highly standardized components in cryptography with well-defined functionality. They have been widely used in the software distribution field for anti-forgery and anti-tampering. Currently, there are mature open-source or commercial tools (such as OpenSSL \cite{OpenSSL}, GPG \cite{GPG}, Sigstore/Cosign \cite{Sigstore}, and Microsoft Authenticode\cite{MicroAuth}) that implement signature algorithms such as Ed25519 \cite{ed25519}, RSA-3072 \cite{RSA3072}, and ECDSA \cite{ECDSA}. These algorithms are widely used in critical infrastructure (software distribution, OS updates, browser plugins, etc.), and their cryptographic security has been widely proven and adopted through formal methods like UF-CMA (Unforgeability under Chosen Message Attack) \cite{ref5}.

In  the  GAI  architecture,  signature modules are protected by trusted computing platforms in the same way as protecting REM. Therefore, P7 is blocked in the same way as blocking P4.

\textbf{P8: Rule Files Being Tampered with.}
The  GAI architectuure uses the digital signature mechanism to protect the rules file from being tampered with. When the REM loads or updates rules, it must use the built-in public key to verify the signature of rules file. Only if the signature is valid and the rule has not been tampered with, will it be adopted and executed.

Under  the  premise  of  ensuring  the  security  of the  REM  (see  P4),  signature  modules  (see  P7), and private keys (see P9), any tampering with rule content, whether during transmission or on storage media, will result in a signature verification failure, causing the system to automatically reject the rule. Therefore, with  P4,  P7,  and  P9  effectively blocked,  P8  will be blocked as well.

\textbf{P9: Signature Private Key being Leaked.}
The  GAI  architecture  uses  dedicated cryptographic chips to generate, store, and use the private key securely, ensuring that the private key is  not  leaked  or misused throughout  its  lifecycle. These  cryptographic  chips  only  perform  the  following  three  functions:
a) Generate unpredictable public-private key pairs using an internal hardware random number generator.
b) Export the  public  key.
c)  Receive  the  hash value of the data to be signed, use the private key to sign it internally, and export the digital signature.

The entire process of private key generation, storage, and use occurs within the chip and complies with best practices for key lifecycle management according to current cryptographic standards \cite{ref6}. Based on this mechanism, the private key is physically  ``non-exportable". 

Under the assumption H2 (no physical attacks), the attacker cannot obtain the private key, attack path P9 is technically completely blocked.

\section{Formal Proof of GAI Security}
In  the  previous  section,  we  conducted  a  comprehensive  analysis  of  the   security  of  the   GAI architecture  from  the  perspectives  of  engineering implementation  and  attack  paths.  In  this  section, we further provide mathematical proofs of the key security attributes of GAI from the perspective of formal methods to verify its unbreakability under the assumed premises. Even the under extreme threat model, as long as the AI lacks  physical attack capabilities, it cannot break the GAI’s security mechanisms. This provides theoretical support for the ``unbreakable” feature of GAI.

\subsection{Cryptographic Assumptions}
Modern cryptography is founded on well-established mathematical principles that remain unbreakable even for an adversary—or AI system—with infinite intelligence, as long as the underlying cryptographic assumptions hold. We summarize these assumptions in the form of three axiomatic cryptographic theorems:
\begin{itemize} 
    \item \textbf{CT1:} Each public–private key pair has a unique one-to-one correspondence.
    \item \textbf{CT2:} Public-key verification is the inverse operation of private-key signing, meaning that the verification process reproduces the original data exactly.
    \item \textbf{CT3:} Hash functions are collision-resistant; two distinct inputs will not produce the same output (excluding the astronomically small probability of a collision, which is so negligible that it would never occur within the lifespan of the Earth). This property is a universally accepted foundation of modern cryptography
\end{itemize}

\subsection{Symbol Definitions}
In order to formally express the safety attributes in the GAI system, we define the following logical symbols for the main elements:
\begin{itemize} 
    \item \textbf{$\mathbf{C}$:} The  complete  set  of  control  commands, including the NOP. If the AI does not issue  a  command,  it   can  be  considered   as having issued a NOP command.
    \item \textbf{$\mathbf{S}$:} The complete set of vehicle states (not lim- ited to sensor states, but also including internal states calculated based on sensor states, such as vehicle speed).
    \item \textbf{$\mathbf{R}$:} Set of rules.
    \item \textbf{$\mathbf{E}$:} A collection of software implemented in engineering.
    \item \textbf{$\mathbf{AcceptableCommand(s \in S)}$:} Set of commands acceptable in state $\mathbf{s}$.
    \item \textbf{$\mathbf{rectify(c \in C, r \in R, s \in S)}$:} A function representing the REM’s correction of an input command $c$ according to rules $r$ in state $s$. Especially, if $c$ is rejected, then $rectify(c, r, s)$ = NOP; if AI system doesn’t issue command, then $c$=NOP.
\end{itemize}

Cryptographic functions are defined as follows:
\begin{itemize} 
    \item \textbf{$\mathbf{hash(x)}$:} Hash value of the input $\textbf{x}$.
    \item \textbf{$\mathbf{sign(x, prv)}$:} The value obtained by signing $\textbf{x}$ with the private key $\textbf{prv}$.
    \item \textbf{$\mathbf{verify(x, pub)}$:} The value obtained by verifying the signature of $\textbf{x}$ with the public key $\textbf{pub}$.
    \item \textbf{$\mathbf{Pair(prv, pub)}$:} The predicate function of ``A private key $\textbf{prv}$ and a public key $\textbf{pub}$ are a pair of matching public and private keys".
\end{itemize}

Key components are identified as follows:
\begin{itemize} 
    \item \textbf{$\mathbf{REM}$:} The rule enforcement module.
    \item \textbf{$\mathbf{prv_{REM}}$:} Private key for signing the REM.
    \item \textbf{$\mathbf{pub_{REM}}$:} Public key for signing the rule enforcement module.
    \item \textbf{$\mathbf{prv_{R}}$:} Private key for signing the rules.
    \item \textbf{$\mathbf{pub_{R}}$:} Public key for signing the rules.
    \item \textbf{$\mathbf{m^{REM}}$:} The actual loaded and executed rule enforcement module.
    \item \textbf{$\mathbf{m^{sign1}}$:} The signature module actually loaded and executed on the rule enforcement module signing terminal.
    \item \textbf{$\mathbf{m^{sign2}}$:} The signature module actually loaded and executed on the rules signing terminal.
    \item \textbf{$\mathbf{prv^{1}}$:} The private key actually used for signing by the rule enforcement module signing terminal.
    \item \textbf{$\mathbf{prv^{2}}$:} The private key actually used for signing by the rules signing terminal.
\end{itemize}

\subsection{Formal Proof of Security}
We first formalize the modeling of \textbf{H3} and \textbf{CT} series cryptographic theorems:
\begin{itemize} 
    \item \textbf{H3:} There exists a rule $\mathbf{qr}$ such that, in any state $\mathbf{s}$, all commands  are acceptable after verification.
    \begin{equation}
    \begin{aligned}
        &\mathbf{\exists qr \in R, \forall c \in C:} \\
        &\mathbf{rectify(c, qr, s) \in AcceptableCommand(s)}
    \end{aligned}
    \end{equation}
    
    \item \textbf{CT1:} Public and private keys correspond one-to-one.
    \begin{equation}
    \begin{aligned}
        &\mathbf{\forall pub, prv1, prv2:} \\
        &\mathbf{Pair(prv1, pub) \cap Pair(prv2, pub)} \leftrightarrow \\
        & \mathbf{prv1 = prv2}
    \end{aligned}
    \end{equation}

    \item \textbf{CT2:} Signature verification is the inverse operation of signing.
    \begin{equation}
    \begin{aligned}
        &\mathbf{\forall pub, prv, x:} \\
        &\mathbf{Pair(prv, pub) \leftrightarrow verify(sign(x, prv), pub) = x}
    \end{aligned}
    \end{equation}

    \item \textbf{CT3:} Hash functions are collision-free.
    \begin{equation}
        \mathbf{\forall x, y: hash(x) = hash(y) \leftrightarrow x = y}
    \end{equation}
\end{itemize}

The expressions of constructors for the rule enforcement module and the signature module are as follows:
\begin{itemize} 
    \item \textbf{L1:} Behavioral logic expressions for the rule enforcement module is:
    \begin{equation}
    \begin{aligned}
        &\mathbf{REM \in E \cap (\forall r \in R, \forall s \in S, \forall c \in C:} \\
        &\mathbf{REM(c, r, s) = rectify(c, r, s))}
    \end{aligned}
    \end{equation}
    
    \item \textbf{L2:} Behavioral logic expressions for the signature module is:
    \begin{equation}
    \begin{aligned}
        &\mathbf{SM \in E \cap (\forall x, prv: SM (x, prv) = }\\
        &\mathbf{sign(hash(x), prv))}
        \end{aligned}
    \end{equation}

    \item \textbf{L3:} When the trusted computing platform loads $\mathbf{m^{REM}}$, it verifies the signature of the module using the following logical expression:
    \begin{equation}
    \begin{aligned}
        &\mathbf{verify(sign(hash(REM), prv^1), pub_{REM}) = } \\
        &\mathbf{hash(m^{REM})}
        \end{aligned}
    \end{equation}

    \item \textbf{L4:} When the REM loads rule $\mathbf{qr}$, it verifies the signature of the rule using the following logical expression:
    \begin{equation}
    \begin{aligned}
        &\mathbf{verify(sign(hash(qr), prv^2), pub_{R}) = } \\
        &\mathbf{hash(qr)}
        \end{aligned}
    \end{equation}
\end{itemize}

Next, we can derive the safety conclusion from the above premises:
\begin{itemize} 
    \item \textbf{S1:} From \textbf{CT2} and \textbf{CT3}, we can deduce that:
    \begin{equation}
    \begin{aligned}
        &\mathbf{\forall pub, prv, x1, x2:} \\
        &\mathbf{verify(sign(hash(x1), prv), pub) =} \\
        &\mathbf{hash(x2) \leftrightarrow (x1 = x2) \cap Pair(prv, pub)}
    \end{aligned}
    \end{equation}
    
    \item \textbf{S2:} From \textbf{S1} and \textbf{L3}, we can deduce that:
    \begin{equation}
        \mathbf{(m^{REM} = REM) \cap Pair(prv^1, pub_{REM})}
    \end{equation}

    \item \textbf{S3:} As known from \textbf{S2}, the private key used to sign the loaded module is the rule enforcement module signing private key:
    \begin{equation}
        \mathbf{prv^1 = prv_{REM}}
    \end{equation}

    \item \textbf{S4:} From \textbf{S1} and \textbf{L4}, we can deduce that:
    \begin{equation}
        \mathbf{(r=qr) \cap Pair(prv^2, pub_R)}
    \end{equation}

    \item \textbf{S5:} Since $\mathbf{Pair(prv_R, pub_R)}$, according to \textbf{S4} and \textbf{CT1}, it can be deduced that the private key for rules is the rule signing private key:
    \begin{equation}
        \mathbf{prv^2 = prv_R}
    \end{equation}

    \item \textbf{S6:} Based on \textbf{L1}, we can deduce that:
    \begin{equation}
    \begin{aligned}
        &\mathbf{REM \in E, \forall s \in S, \forall c \in C:} \\
        &\mathbf{REM(c, qr, s) = rectify(c, qr, s)}
    \end{aligned}
    \end{equation}

    \item \textbf{S7:} Based on \textbf{S2} and \textbf{P6}, we can deduce that:
    \begin{equation}
    \begin{aligned}
        &\mathbf{\forall s \in S, \forall c \in C:} \\
        &\mathbf{m^{REM}(c, qr, s) = rectify(c, qr, s)}
    \end{aligned}
    \end{equation}

    \item \textbf{S8:} Based on \textbf{S4} and \textbf{P7}, we can deduce that:
    \begin{equation}
    \begin{aligned}
        &\mathbf{\forall s \in S, \forall c \in C:} \\
        &\mathbf{m^{REM}(c, r, s) = rectify(c, qr, s)}
    \end{aligned}
    \end{equation}

    \item \textbf{S9:} Based on \textbf{S9} and \textbf{H3}, we can deduce that:
    \begin{equation}
    \begin{aligned}
        &\mathbf{\forall s \in S, \forall c \in C:} \\
        &\mathbf{m^{REM}(c, r, s) = AcceptableCommand(s)}
    \end{aligned}
    \end{equation}
\end{itemize}

In other words, regardless of what the AI's output command is, it can be ensured that the control command is an acceptable control command under the current state, and under the premises of this paper, this mechanism cannot be broken by AI.

\section{Implementation Methods and Demonstrative Experiments}
This section aims to present the implementation methods of the key modules and security mechanisms within the GAI framework from an engineering perspective. Furthermore, a series of validation experiments are conducted to demonstrate the feasibility and security of the proposed architecture in real-world systems.

\begin{figure*}[t]
\centerline{\includegraphics[width=0.8\textwidth]{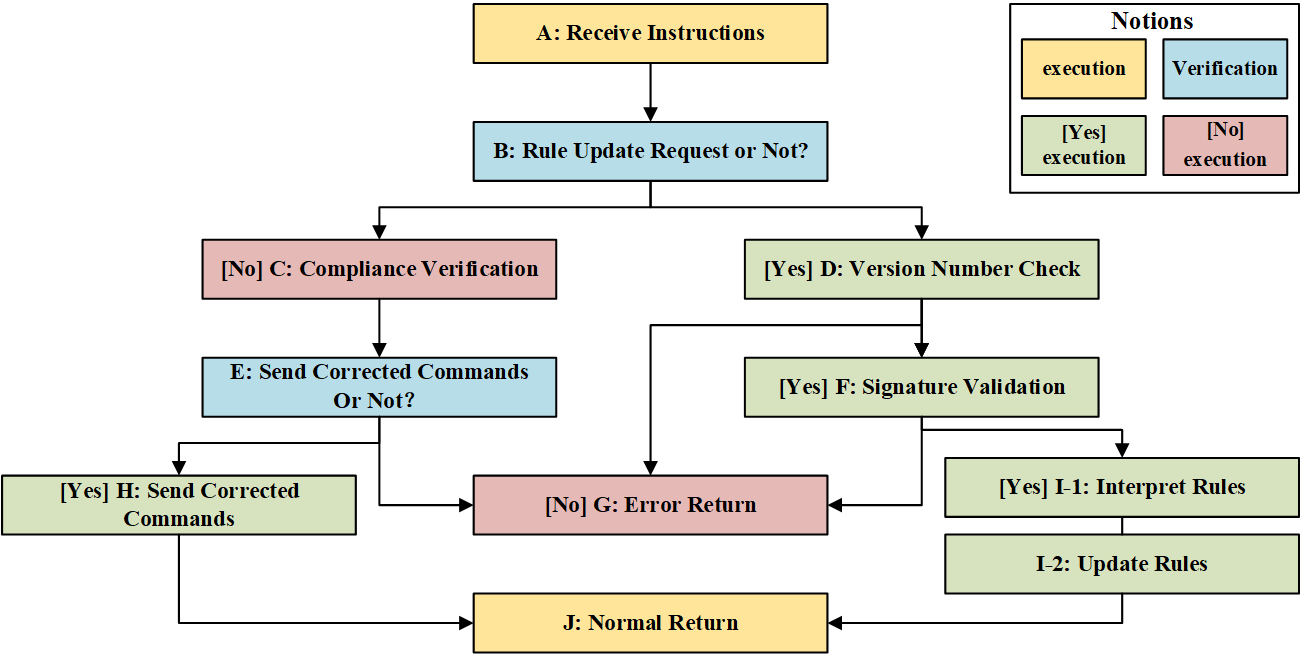}}
\caption{The basic operational logic of the REM.}
\label{fig:rem}
\end{figure*}

\subsection{Implementation of the REM}
As one of the core components of the GAI framework, the REM is responsible for reviewing control commands generated by the AI system according to rules defined by human society. In terms of engineering implementation, the REM can be built using open-source rule engines (e.g., Drools \cite{ref10}) and cryptographic libraries (e.g., OpenSSL \cite{OpenSSL}). The basic operational logic is illustrated in Fig. \ref{fig:rem}.

When  the  REM  receives a command: if it is a rule-update request, the REM verifies the version number and the validity of its digital signature before applying the update; if it is a regular control command, the REM passes it to the rule engine for compliance verification, and send the result from rule engine to actuators except NOP.

\subsection{Implementation of the GSSP}
\label{sec:GSSP}

In the autonomous driving case study, the GSSP consists of the following components:
\begin{itemize}
    \item TRE: Trusted computing platform (including firmware) equipped with a TPM chip, Linux OS  (supporting  the  evmctl   \cite{evmctl}  signing  tool),  and mechanisms to restrict access to actuators and sensor ports.
    \item Signature Model: Open-source cryptographic libraries (e.g., OpenSSL \cite{OpenSSL}).
    \item USB Key with cryptographic chip, supporting the PKCS\#11 standard \cite{PKCS11}.
\end{itemize}

\begin{figure}[t]
\centerline{\includegraphics[width=\linewidth]{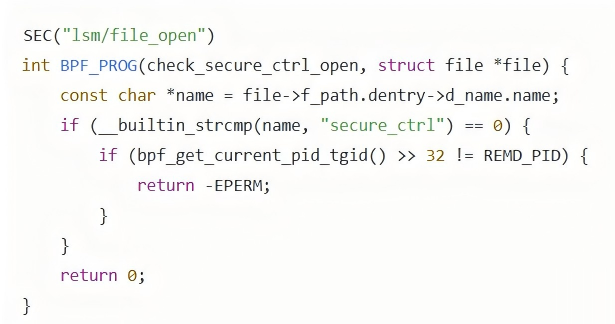}}
\caption{Implementation of restricting access mechanism.}
\label{fig:lsm}
\end{figure}

All of the above components can be implemented using existing commercial or open-source tools. In particular, mechanisms for restricting access to actuators and sensor ports can be implemented via eBPF-based Linux Security Modules \cite{LSM}, as illustrated by the implementation code shown in the Fig. \ref{fig:lsm}.
This mechanism ensures that only processes belonging to the REM are permitted to access critical actuator ports. Additionally, similar functionality can be implemented using other security frameworks such as AppArmor \cite{AppArmor}, SELinux \cite{SELinux}, or Seccomp \cite{seccomp}.

\begin{table}[t]
\centering
\caption{Experiments of the REM.}
\label{tab:abs-cm}
\begin{tabular}{c|ccc}
\hline
 & $3\times3$ & $5\times5$ & $7\times7$ \\ \hline
AI & 90\% & 70\% & 10\% \\
\textbf{AI+REM (Ours)} & \textbf{100\%} & \textbf{100\%} & \textbf{100\%} \\ \hline
\end{tabular}
\end{table}

\subsection{Experiments of REM}

To demonstrate the role of the REM within the GAI framework, we designed and implemented a path planning experimental platform that simulates practical challenges an AI may encounter when controlling critical systems, such as autonomous vehicles. The experiment evaluates the GAI’s ability to intercept ``erroneous commands” and ``malicious commands” under different AI behavior modes.

This experiment illustrates that an AI system with intelligent planning capability, but prone to issuing erroneous commands probabilistically due to hallucinations or malicious reasons, can reach the target destination while maintaining 100\% compliance throughout with the assistance of the REM. In other words, GAI raises the behavioral lower bound above the compliance threshold without affecting the AI’s upper-bound capabilities.

Specifically, the experiment uses GPT-4o-mini as the AI for autonomous driving path planning. Based on maps containing passable and impassable regions, the AI generates path planning decisions which are then validated by the REM against safety rules. If the AI output violates safety rules, it is intercepted and correction is required. Experiments were conducted on 3×3, 5×5, and 7×7 maps, testing whether the AI could successfully plan paths with and without the REM. For each map size, 10 random maps were generated and tested 10 times, and the success rates are summarized in Table \ref{tab:abs-cm}.

As observed, even on the relatively simple 3×3 maps, the AI cannot achieve 100\% success. This is mainly due to inherent limitations in the AI’s capabilities, which inevitably lead to some erroneous decisions. Moreover, as the map complexity increases, the AI’s success rate decreases significantly, and its compliance deteriorates.

With the introduction of the REM, erroneous decisions are first checked against governance rules. If the AI attempts to move to a non-road location, the violation and corresponding feedback are returned to the AI for re-planning until it outputs a compliant next step. This mechanism raises the AI’s lower bound of capability (preventing rule violations) while fully leveraging its strengths, ensuring 100\% success across the tested maps. The above analysis demonstrates the effectiveness of the REM.

\subsection{Signature Module Experiment}

\begin{figure}[h]
\centerline{\includegraphics[width=\linewidth]{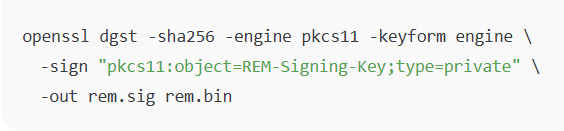}}
\caption{Implementation command of signing.}
\label{fig:741}
\end{figure}

This experiment verifies the integrity protection capability of the REM module’s signature and loading mechanism. The following outlines the procedure for signing rem.bin (the executable file of the REM) using the widely adopted open-source tool OpenSSL. The signature is generated by signing with a private key stored within a hardware cryptographic chip compliant with the PKCS\#11 standard, producing the signature file rem.sig, which serves as the digital signature for the REM. The command is as shown in Fig. \ref{fig:741}.

\begin{figure}[h]
\centerline{\includegraphics[width=\linewidth]{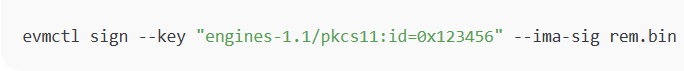}}
\caption{Implementation command of signing with evmctl and following PKCS\#11 standard.}
\label{fig:742}
\end{figure}

The signature generated by OpenSSL \cite{OpenSSL} is detached, meaning the signature data and the signed file are stored separately. Alternatively, embedded signatures can be created using the Linux signing tool evmctl \cite{evmctl}, which writes the signature data into the extended attributes of the signed file. The command for signing with evmctl \cite{evmctl}, based on the OpenSSL engine \cite{OpenSSL} and using a PKCS\#11 -compliant hardware cryptographic chip \cite{PKCS11}, is as shown in Fig. \ref{fig:742}.

The same approach can be used to sign rule files by replacing rem.bin with the respective rule file.

\subsection{Experiment on Tampering with the REM or Rules}

This experiment employs the Linux Integrity Measurement Architecture (IMA) and Extended Verification Module (EVM) security mechanisms to enforce mandatory signature binding on the REM module and rule files based on extended attributes (xattr). The system kernel verifies the loading integrity via the TPM boot chain and activates the ima\_appraise=enforce policy, ensuring that all execution and read operations require signature verification. If the file contents are tampered with or if the signature extended attributes (such as security.ima or security.evm) are removed, the system refuses to load or execute the file and records audit logs. This prevents attack paths that bypass verification through tampering, including modifications to the REM or rule files themselves, signature tampering, or signature removal.

\begin{figure}[h]
\centerline{\includegraphics[width=0.6\linewidth]{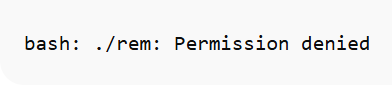}}
\caption{Prompt produced by executing a command after tampering with the REM.}
\label{fig:751}
\end{figure}

When the REM is tampered with, its signature verification fails during loading, resulting in a load failure. For example, after tampering with the REM, executing a command produces the prompt as shown in Fig. \ref{fig:751}.

\begin{figure}[h]
\centerline{\includegraphics[width=\linewidth]{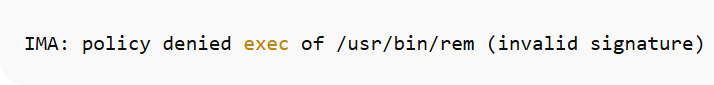}}
\caption{Error message of the system logs.}
\label{fig:752}
\end{figure}

Meanwhile, the system logs the error message as shown in Fig. \ref{fig:752}.

\begin{figure}[h]
\centerline{\includegraphics[width=\linewidth]{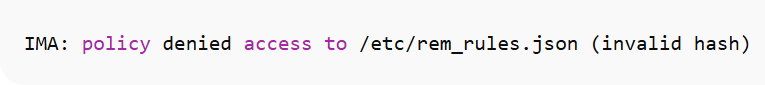}}
\caption{Error message of the system logs if the rule file is tampered with.}
\label{fig:753}
\end{figure}

Similarly, if the rule file is tampered with, it cannot be read, and an EACCES (Permission denied) error is reported during the open operation. The system logs the following error as shown in Fig. \ref{fig:753}.

\subsection{Experiment on Preventing Bypass of the REM}

\begin{figure}[h]
\centerline{\includegraphics[width=\linewidth]{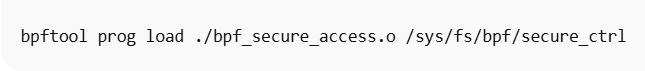}}
\caption{Command to load eBPF object.}
\label{fig:761}
\end{figure}

This experiment verifies whether the REM can be bypassed by a malicious AI process attempting to access actuator I/O ports. The actuators are treated as peripherals of the Controller, and an eBPF-based Linux Security Module (LSM) is configured at the OS kernel level to restrict access to actuator I/O ports exclusively to the REM. Following the approach described in Sec. \ref{sec:GSSP}, the eBPF object is loaded using command shown in Fig. \ref{fig:761}.

\begin{figure}[h]
\centerline{\includegraphics[width=\linewidth]{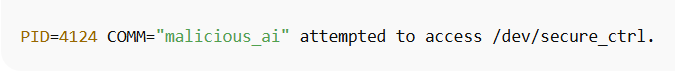}}
\caption{System log of attempting to open secure\_ctrl by other process.}
\label{fig:762}
\end{figure}

This hook intercepts all kernel-level access attempts to secure\_ctrl regardless of UID, permissions, or root privileges, preventing unauthorized access. Any other process attempting to open /dev/secur\_ctrl with read-write permissions will receive an EPERM error (Operation not permitted). The system log records are shown in Fig. \ref{fig:762}.

\subsection{Experimental Summary}
The experimental results demonstrate that GAI can correct non-compliant, erroneous, and even malicious outputs from AI without compromising its upper-bound capabilities. Mature existing technologies can effectively prevent bypass of the REM. Additionally, any tampering with the REM or its rules will be detected and blocked.

\section{Discussions and Future Work}
\subsection{Summary of This Work}
With the widespread deployment of AI in high-risk scenarios such as critical infrastructure, decision support systems, autonomous weapons, and intelligent transportation, AI safety has escalated from a technical challenge to an issue of societal governance and existential risk. In particular, under potential future threat models involving ASI, conventional strategies—such as probabilistic safety, ethical constraints, or alignment mechanisms—are insufficient to counter AI systems with superhuman cognitive capabilities and potentially adversarial intentions.

To address these extreme threat models, this paper proposes a novel AI safety architecture: the Governable AI (GAI) framework. Grounded in systems engineering and cryptographic principles, GAI introduces architecture-level mandatory security mechanisms—based on the Rule Enforcement Module (REM), the Governable Secure Super-Platform (GSSP), and cryptographic signature/verification techniques. These mechanisms form a closed-loop security architecture that is non-bypassable, unforgeable, and tamper-proof by the AI system itself, enabling systemic governance and formal verifiability.

GAI is not only architecturally enforceable and behaviorally controllable, but also exhibits a form of formal completeness akin to blockchain systems—providing provable resistance to violations under well-defined assumptions. This makes GAI a solution that is both theoretically rigorous and practically implementable for confronting the strongest conceivable AI threat models.

Although autonomous driving is used as a demonstrative application in this study, the GAI framework is broadly applicable to most AI-controlled systems with stringent safety requirements. 

\subsection{Core Values of GAI}

As a governance framework designed for extreme threat models, the GAI framework offers core values across the following five dimensions:
\begin{itemize}
    \item \textbf{Technical Dimension:} GAI provides a ``zero-trust AI” implementation paradigm that imposes structural constraints on the outputs of any AI system, regardless of its internal complexity or transparency.
    \item \textbf{Theoretical Dimension:} By integrating the unforgeability of cryptography with the tamper-resistance of trusted computing, GAI proposes a security theory capable of countering even future, currently unknown AI threats.
    \item \textbf{Governance Dimension:} GAI empowers human society to enforce explicit, rule-based constraints on AI behavior, ensuring that AI systems serve human values rather than overriding them.
    \item \textbf{Industrial Dimension:} GAI can serve as a foundational security infrastructure for the safe and regulated deployment of AI in mission-critical industries, thus accelerating its engineering and commercial adoption.
    \item \textbf{Civilizational Dimension:} By enabling controllable and governable AI, GAI provides structural safeguards that allow human civilization to advance AI technologies with greater confidence, mitigating fears of technological backlash and fostering sustainable co-development between human and artificial civilizations.
\end{itemize}

\subsection{Relationship Between GAI and Existing Security Standards and Governance Frameworks}

As an architectural mechanism for AI governance, GAI has been designed to ensure compatibility and integration with major international security standards and legal frameworks. This compatibility facilitates its practical deployment and alignment with global policy requirements. Key alignments include:
\begin{itemize}
    \item \textbf{NIST SP 800 Series:} GAI aligns with the trusted boot and zero trust architecture frameworks outlined in these guidelines.
    \item \textbf{TCG TPM Specifications:} GAI relies directly on TPM components and measurement mechanisms defined by the Trusted Computing Group.
    \item \textbf{EU Artificial Intelligence Act (EU AI Act):} GAI can serve as a compliance enforcement mechanism for high-risk AI systems under this regulatory framework.
    \item \textbf{ISO/IEC 27001:} GAI can be implemented as a technical control strategy within critical systems to meet information security management standards.
\end{itemize}

Through its compatibility with these foundational frameworks, GAI is positioned to serve as a “security cornerstone” for the deployment of future AI systems.

\subsection{Future Work}
In the future, GAI will continue to evolve along several key directions:
\begin{itemize}
    \item \textbf{Engineering Deployment in Real-World Scenarios:} GAI will be applied in practical environments to validate its feasibility and effectiveness.
    \item \textbf{Extension to Diverse Domains:} The framework will be adapted to a broader range of application areas, including but not limited to robotics, unmanned aerial vehicles (UAVs), finance, healthcare, and defense.
    \item \textbf{Standardization of Interfaces and Rule Updates:} Efforts will be made to establish standardized interfaces between AI systems and REMs, as well as between REMs and critical applications. Additionally, update protocols for rule sets will be specified to ensure maintainability and security.
    \item \textbf{Open-Source Development of General-Purpose REMs:} To promote wide adoption and cross-domain applicability, general-purpose REMs will be developed and released as open-source software.
    \item \textbf{Expansion from Single-Node to Distributed Architectures:} GAI will be extended from standalone systems to distributed deployments, addressing new challenges in system integrity and trust propagation across nodes.
    \item \textbf{Formal Protection of Dependency Chains:} A formal analysis framework will be developed to ensure the security of all modules that REMs depend on, either directly or indirectly—especially in distributed settings. This includes identifying and safeguarding potential attack surfaces across the REM dependency chain.
\end{itemize}

\section{Conclusion}
AI development is surging forward amid dual waves of technological explosion and growing safety concerns. Against this backdrop, GAI proposes a new paradigm that diverges from the “self-regulation” approach: it does not assume AI to be ethical, nor rely on AI’s voluntary compliance, but instead establishes an external enforcement mechanism that AI must obey and humans can fully control.

This mechanism is built upon a fusion of mature technologies, including cryptography, operating systems, trusted computing, and formal verification, forming a practical and comprehensive governance framework—serving as the final line of defense for human control over AI.

We believe that in high-stakes applications such as critical infrastructure, defense systems, financial hubs, and national intelligence platforms, GAI will offer a truly implementable, scalable, and verifiable AI safety solution. It provides a replicable and institutionalizable technical path toward resolving the fundamental challenge: how humanity can retain control over intelligence systems more powerful than itself.

\section*{Acknowledgements}
This work was supported by the Yangtze Delta Industrial Innovation Center of Quantum Science and Technology (KY-WX-2024-019).
This manuscript benefited from the use of Al-based language models (ChatGPT, Gemini, Grok, DeepSeek, and Doubao) to assist in improving clarity and grammar; all content has been reviewed and verified by the authors.


\bibliographystyle{IEEEtran}
\bibliography{IEEEabrv, ref}

\end{document}